# DCFG: Diverse Cross-Channel Fine-Grained Feature Learning and Progressive Fusion Siamese Tracker for Thermal Infrared Target Tracking

Ruoyan Xiong[1,2][†], Yuke Hou[1,2][†], Princess Retor Torboh[2][*], Hui He[3], Huanbin Zhang[1,2], Yue Zhang[1,2], Yanpin Wang[1,2], Huipan Guan[1,2], and Shang Zhang[1,2]

[1] College of Computer and Information Technology, China Three Gorges University, Yichang 443002, China
[2] Hubei Province Engineering Technology Research Center for Construction Quality Testing Equipment, China Three Gorges University, Yichang 443002, China
[3] Wuhan College, Wuhan 430212, China
torbohprincessretor@gmail.com

**Abstract.** To address the challenge of capturing highly discriminative features in thermal infrared (TIR) tracking, we propose a novel Siamese tracker based on cross-channel fine-grained feature learning and progressive fusion. First, we introduce a cross-channel fine-grained feature learning network that employs masks and suppression coefficients to suppress dominant target features, enabling the tracker to capture more detailed and subtle information. The network employs a channel rearrangement mechanism to enhance efficient information flow, coupled with channel equalization to reduce parameter count. Additionally, we incorporate layer-by-layer combination units for effective feature extraction and fusion, thereby minimizing parameter redundancy and computational complexity. The network further employs feature redirection and channel shuffling strategies to better integrate fine-grained details. Second, we propose a specialized cross-channel fine-grained loss function designed to guide feature groups toward distinct discriminative regions of the target, thus improving overall target representation. This loss function includes an inter-channel loss term that promotes orthogonality between channels, maximizing feature diversity and facilitating finer detail capture. Extensive experiments demonstrate that our proposed tracker achieves the highest accuracy, scoring 0.81 on the VOT-TIR 2015 and 0.78 on the VOT-TIR 2017 benchmark, while also outperforming other methods across all evaluation metrics on the LSOTB-TIR and PTB-TIR benchmarks.

**Keywords:** Thermal Infrared Target Tracking, Fine-grained Features, Siamese Network, Progressive Feature Fusion.

## 1 Introduction

Thermal infrared (TIR) target tracking is a critical area of research in computer vision. Due to the ability of infrared sensors to penetrate fog and their independence from ambient lighting, TIR tracking is particularly effective in challenging environments, such as rain, fog, low illumination, and intense lighting. These characteristics enable TIR tracking systems to operate reliably in various weather conditions, offering strong anti-interference capabilities. As a result, TIR tracking is widely applied in fields such as autonomous driving, maritime rescue, and weapon guidance. However, TIR targets in infrared imagery often suffer from poor resolution, low signal-to-noise ratios, weak contrast, and lack of texture details. Additionally, changes in target appearance and scene complexity, such as occlusion, scale variations, thermal crossover, and background clutter, further complicate the differentiation between targets and background. Therefore, achieving robust TIR target tracking remains a highly challenging task.

---

[†]These authors contributed equally to this work.
[*]Corresponding author.

Siamese network-based trackers first establish a template to represent the target and then select the candidate in subsequent frames that most closely matches the template as the tracking result. Common representation models include raw intensity features, intensity histograms, and gradient features. For instance, Cheng et al. [1] constructed a cascaded grayscale space and used two orthogonal filters to extract directional subspace features for target representation. Paravati et al. [2] employed a genetic algorithm that simulates gene mutation and natural selection, directly using the raw intensity features and computing similarity between candidates and the template via cross-correlation.

Building upon advances in visual tracking, some TIR trackers adopt deep features for target representation. For example, Gao et al. [3] proposed a large-margin structured convolutional operator that integrates structured output support vector machines and discriminative correlation filters. This operator generates continuous appearance and motion feature maps through spatial regularization and implicit interpolation, and updates the operator via joint optimization. Zulkifley et al. [4] introduced a multi-model fully convolutional neural network method, which generates candidate target locations through a two-stage sampling process. It selects the best sample based on appearance similarity, predicted location, and model reliability, updates the appearance model with accumulated positive and negative samples, and adaptively adjusts the sampling variance based on tracker confidence. Similarly, Wu et al. [5] incorporated a feature attention module and an expanded search strategy into a fully convolutional classifier, using a pretrained residual network to link two convolutional layers and regress the confidence of the target location online. Li et al. [6] proposed a mask sparse representation model that combines sparse coding with high-level semantic features. However, relying solely on deep features can lead to tracking failures in some cases, as similar TIR targets often share highly similar visual and semantic characteristics. To address this limitation, some studies have attempted to learn fine-grained TIR features, such as local structures and contours, to extract more discriminative representations. For example, Sun et al. [7] suppressed salient features to force the network to rely on subtle cues for classification, optimizing the classification of confusing categories with a gradient boosting loss. Chang et al. [8] introduced an inter-channel loss and a channel-level attention mechanism to enhance the discriminability of features within the same category and ensure spatial exclusivity among feature channels, thus improving classification performance. Wang et al. [9] proposed a hierarchical pyramid diversity attention network to capture multi-scale features, reduce feature redundancy, and fuse multi-level information using hierarchical bilinear pooling.

Unlike previous methods, this paper handles a critical limitation in existing networks, which often focus only on the most prominent parts of the target due to insufficient constraints, thereby failing to capture a broader range of spatial regions. As a result, subtle features that help differentiate the target from background interference are often overlooked. To overcome this issue, we propose DCFG, a novel Siamese tracker based on diverse cross-channel fine-grained feature learning and progressive fusion. In particular, we design a diverse cross-channel fine-grained feature learning network that extracts detailed TIR target representations. By applying feature masks and suppression coefficients, the network downweights prominent features, encouraging the tracker to capture richer fine-grained cues. A channel reshuffling mechanism is introduced to maintain effective information flow, while channel balancing reduces parameter count and enhances model efficiency. Additionally, a layer-wise combination unit progressively extracts and fuses features, minimizing parameter overhead and computational cost. To further improve feature fusion, a progressive fusion module integrates both feature redirection and channel shuffle operations. Since fine-grained features often concentrate in localized regions, relying solely on these mechanisms may not fully capture the diversity of target characteristics. To address this, we introduce a diverse cross-channel fine-grained loss function, which guides the network to discover subtle features of TIR targets, thereby enhancing target representation. This loss function includes an inter-channel term that enforces orthogonality of features across channels, maximizing diversity and enabling the tracker to learn a broader range of fine-grained patterns.

The main contributions of this work are summarized as follows:

- We propose a novel Siamese tracker based on diverse cross-channel fine-grained feature learning and progressive fusion, addressing the limitation of focusing solely on prominent target features.
- We introduce a cross-channel fine-grained feature learning network that uses feature masks and suppression coefficients to downweight dominant features, thereby encouraging the extraction of richer fine-grained cues.
- We propose a loss function that guides different feature groups to focus on distinct discriminative regions of the target, improving target representation.
- Extensive experiments on PTB-TIR, LSOTB-TIR, VOT-TIR2015, and VOT-TIR 2017 benchmarks demonstrate the effectiveness of our tracker for TIR tracking.

## 2  Methodology

### 2.1  Overview of DCFG

Fine-grained features are crucial for distinguishing distractors and maintaining robust TIR tracking. Most existing TIR trackers predominantly focus on the most salient target regions, making them susceptible to visually similar distractors, which can lead to tracking failures [10]. To address this limitation, we propose DCFG tracker, which integrates cross-channel fine-grained feature learning with progressive fusion to enhance target representation capability. As illustrated in Fig. 1, the proposed DCFG comprises two main modules: (a) a backbone network, and (b) a DCFG Region Proposal

Network (DCFG_RPN). The tracking task is formulated as a cross-correlation problem within a Siamese network architecture, which computes similarity maps from deep feature embeddings. The template branch encodes target features, while the search branch processes candidate regions, with both branches sharing convolutional neural network parameters to ensure consistent feature transformation.

As shown in Fig. 1(a), the backbone network of the proposed DCFG is based on a modified ResNet-50 architecture. The original ResNet employs a stride of 32 pixels, which is inadequate for dense sampling of fine-grained features required in Siamese networks. To address this limitation, we reduce the strides of the Conv4 and Conv5 blocks from 16 and 32 pixels to 8 pixels, respectively, and introduce dilated convolutions to expand the receptive field, thereby capturing richer contextual information. Additionally, a 1×1 convolutional layer is appended to each residual block output, reducing the channel dimension to 256. To support joint inference of the location of the target, we further extract features from multiple branches. Specifically, outputs from Conv3, Conv4, and Conv5 are individually fed into three separate DCFG_RPN modules, facilitating hierarchical aggregation of multi-level features.

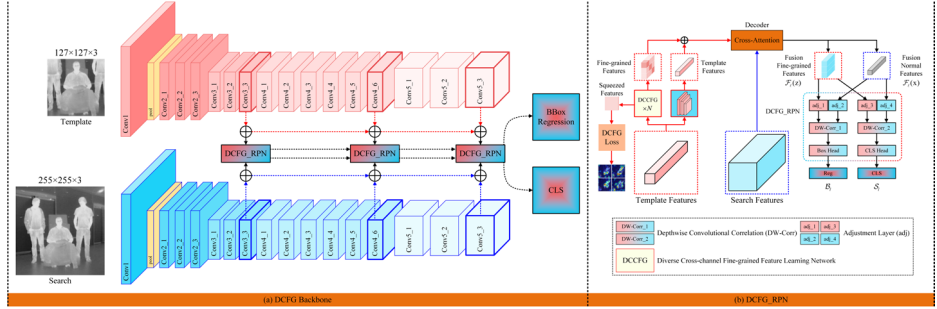

**Fig. 1.** The architecture of the proposed DCFG.

As illustrated in Fig. 1(b), each DCFG_RPN module begins by extracting features from the template and search regions using a CNN and encoder. These features are then processed through a diverse cross-channel fine-grained feature learning network to obtain detailed target representations, after which the encoder fuses these outputs. Furthermore, a depth-wise cross-correlation operation (DW-Corr) performs channel-wise correlation, effectively capturing spatial characteristics within each input channel. Finally, the resulting correlation maps are passed through a head module consisting of a cross-correlation layer and fully convolutional layers, generating classification (CLS) scores and bounding box regression (Reg) values. All outputs are integrated using weighted fusion, jointly estimating the final target position.

### 2.2 Diverse Cross-Channel Fine-Grained Feature Learning Network

Existing TIR tracker typically focus on the most salient target regions, neglecting subtle features essential for distinguishing targets from distractors. To effectively capture discriminative TIR features, we propose a Diverse Cross-Channel Fine-Grained Feature Learning (DCCFG) Network. In this network, template features are partitioned into multiple channel groups, each encoding distinct fine-grained representations.

Specifically, the proposed DCCFG consists of two parallel branches, mask suppression and weight suppression, forming a Siamese structure as illustrated in Fig. 2(a). DCCFG is composed of four key components. First, a channel rearrangement module is introduced to facilitate sufficient information flow and moderate feature interaction between channels. Second, channel equalization aligns the number of input and output channels, thereby reducing the total number of parameters and improving model efficiency. Third, a layer-wise fusion unit progressively extracts and merges features, minimizing redundancy and computational cost. Finally, a channel shuffle operation redistributes features across channels to efficiently integrate fine-grained information.

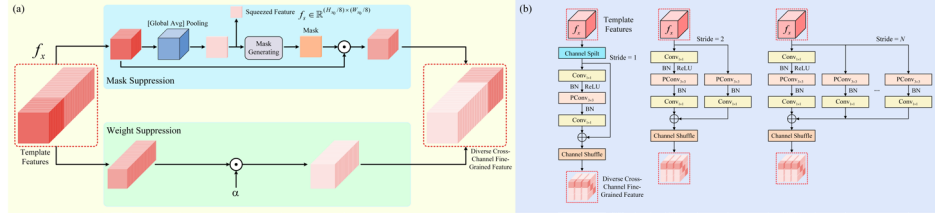

**Fig. 2.** Diverse cross-channel fine-grained feature learning network.

**Backbone Structure.** In the DCCFG network, the mask suppression branch receives the features from the $i$-th group of the template feature map $f_m$, while the weight suppression branch processes the remaining feature groups, denoted as $f_s$. The mask suppression branch first identifies key regions in $f_m$, typically corresponding to peak response locations, and then applies a suppression mask to attenuate those high-activation regions. To identify the peak response position in $f_m$, we employ Global Average Pooling (GAP), which computes the average response across all channels. This results in a compressed feature map $f_s \in \mathbb{R}^{(H_{x_0}/8) \times (W_{x_0}/8)}$. Since $f_s$ represents the mean activation map, the location of its maximum value corresponds to the most salient region in $f_m$, denoted as $(x_{max}, y_{max})$. A suppression mask $M$ is generated to suppress the feature map $f_m$. The suppression mask $M$ is defined as:

$$M_{i,j} = (1 - Gauss(i,j)) \tag{1}$$

where the Gaussian function is defined as:

$$Gauss(i,j) = \frac{1}{2\pi\sigma^2} e^{-\frac{(i-x_{max})^2 + (j-y_{max})^2}{2\sigma^2}} \tag{2}$$

where $i = 1,2,3,\ldots,W_{x_0}/8$, $j = 1,2,3,\ldots,H_{x_0}/8$.

To promote diversity in feature extraction while maintaining the overall feature integrity, a suppression coefficient $\alpha \in [0,1]$ is applied in the weight suppression branch to reduce the contribution of features in $f_w$.

The final suppressed template feature $f_{\hat{x}}$ is defined as:

$$f_{\hat{x}} = \begin{cases} M \odot f_{x_i} & \text{if} \quad f_{x_i} \in f_m \\ \alpha^* f_{x_i} & \text{if} \quad f_{x_i} \in f_w \end{cases} \tag{3}$$

where $\odot$ denotes element-wise multiplication, $i$ represents the feature group index, and the feature map is divided into $N$ groups. According to ablation studies, optimal performance is achieved when $N = 8$ and $\alpha = 0.7$.

**Channel Balance and Lightweight Strategy.** The proposed network adopts depthwise separable convolution, where the pointwise convolution (PWConv) dominates the computational cost [11]. Let $C_{in}$ denote the number of input and output channels, respectively. The floating-point operations (FLOPs) and memory access cost (MAC) of the pointwise convolution are defined as:

$$F = H \times W \times C_{in} \times C_{out} \tag{4}$$

$$MAC = H \times W \times (C_{in} + C_{out}) + C_{in} \times C_{out} \tag{5}$$

where $H$ and $W$ represent the height and width of the feature map.

To reduce MAC, a channel equalization strategy is employed by setting the input-to-output channel ratio approximately to $C_{in}:C_{out} \approx 1:1$. Applying the arithmetic mean inequality yields the MAC lower bound:

$$MAC \geq \sqrt[2]{H \times W \times F} + \frac{F}{H \times W} \tag{6}$$

This implies that MAC reaches its minimum when the number of input and output channels is balanced, effectively reducing the overall computational complexity of the network. Additionally, group convolution is incorporated to convert dense inter-channel operations into sparse ones, further lowering FLOPs. However, using an excessively large number of groups can adversely affect runtime efficiency. Therefore, for pointwise convolution, the MAC and FLOPs with group convolution are given by:

$$\begin{aligned} MAC_{PWConv} &= H \times W \times (C_{in} + C_{out}) + \frac{C_{in} \times C_{out}}{g} \\ &= H \times W \times C_{in} + \frac{F \times g}{C_{in}} + \frac{F}{H \times W} \end{aligned} \tag{7}$$

$$F_{PWConv} = \frac{H \times W \times C_{in} \times C_{out}}{g} \tag{8}$$

where $g$ is the number of groups. In tracking tasks, given a fixed input shape $H \times W \times C_{in}$ and computational budget $F$, increasing $g$ leads to higher MAC. To balance FLOPs reduction with runtime efficiency, we set $g = 8$ in this work, achieving an effective trade-off between convolution simplification and computational cost.

**Cross-Channel Fine-Grained Feature Block.** As shown in Fig. 2(b), cross-channel fine-grained feature block consists of a sequence of operations, including $Conv_{1\times1}$ and $PConv_{3\times3}$ convolutions layers, followed by batch normalization (BN) and ReLU activation functions. To improve the model's ability to capture subtle variations, a random channel shuffle is applied during the process to enhance inter-channel interaction. At the start of each block, the input feature map with $C$ channels is divided into two groups: $C_0$ and $C - C_0$. One group is retained as-is, while the other is passed through three convolution layers with consistent input and output dimensions. Since the split inherently separates the features into independent groups, the $1 \times 1$ convolution layers do not require additional group operations. After convolution, the two branches are concatenated to restore the original channel dimension, followed by a channel shuffle operation that facilitates cross-branch information exchange. ReLU and BN operations are selectively applied only to the transformed branch to maintain computational efficiency and feature diversity. In downsampling scenarios where the stride is greater than 1 (*i.e.*, ranging from 2 to $N$), the channel splitting step is omitted. Instead, additional $Conv_{1\times1}$ and $PConv_{3\times3}$ layers are employed to reduce spatial resolution and preserve inter-channel interactions, ensuring the maintenance of rich feature representations.

## 2.3 DCFG Loss Function

In the absence of effective constraints, individual template feature groups within the fine-grained feature block may fail to focus on distinct regions of the target. Additionally, without strong supervisory signals, the network may lack the capability to attend to diverse discriminative regions. To address these challenges, we propose a diverse loss function that enhances the learning of TIR features by promoting spatial diversity and fine-grained representation. The proposed loss function consists of three components: the tracking loss $L_{CFGB}$ from the Cross-Channel Fine-Grained Block (CFGB), the tracking loss $L_{norm}$, and the DCFG loss $L_{DCFG}$, which encourages the network to extract unique fine-grained features from different channel groups

The total loss function is formulated as:

$$Loss = \sum L_{CFGB}(F_{\tilde{x}}) + \mu \times L_{norm}(f_x) + L_{DCFG}(F_s) \qquad (9)$$

where $F_{\tilde{x}} = (f_{\tilde{x}_1}, f_{\tilde{x}_2}, \ldots, f_{\tilde{x}_N})$ denotes the output feature maps from the CFGB network, $F_s = (f_{s_1}, f_{s_2}, \ldots, f_{s_N})$ represents the corresponding compressed features, $N$ is the number of CFGBs, and $\mu$ is a balancing hyperparameter.

**Tracking Loss Function.** The losses $L_{CFGB}$ and $L_{norm}$ share a common tracking loss function, denoted as $L_{track}$. This function consists of two components: a classification loss $L_{cls}$ for distinguishing the target from the background, and a regression loss $L_{reg}$ for bounding box prediction. The total tracking loss is defined as:

$$L_{track} = L_{cls} + L_{reg} \qquad (10)$$

The classification loss $L_{cls}$ is formulated as:

$$L_{cls} = -\sum_{j} \left[ y_j \times log(p_j) + (1 - y_j) \times log(1 - p_j) \right] \quad (11)$$

where $y_j$ denotes the ground truth label of the $j$-th sample, and $p_j$ represents the predicted probability output by the network. The regression loss $L_{reg}$ is defined as:

$$L_{reg} = \sum_{j} 1_{\{y_i=1\}} \times \left[ \lambda_G \times L_{IoU}(b_j, \hat{b}) + \lambda_1 \times L_1(b_j, \hat{b}) \right] \quad (12)$$

where $L_{IoU}$ is the Intersection over Union (IoU) loss, $L_1(\cdot)$ represents the L1-norm loss, $\lambda_G$ and $\lambda_1$ are regularization parameters, and $b_j$ and $\hat{b}$ represent the predicted and ground truth bounding boxes, respectively. The indicator function $1_{\{y_i=1\}}$ returns 1 if the sample is positive, and 0 otherwise. As shown in Fig. 3, a tracking loss weight $\mu$ is introduced to integrate the cross-channel fine-grained loss $L_{CFGB}$, which supervises both fine-grained and normal features, with the conventional tracking loss $L_{norm}$. This approach allows the network to learn detailed TIR target characteristics while preserving salient features, without altering the underlying architecture. In this work, $\mu = 4$ is chosen to balance the contributions of fine-grained and salient TIR features.

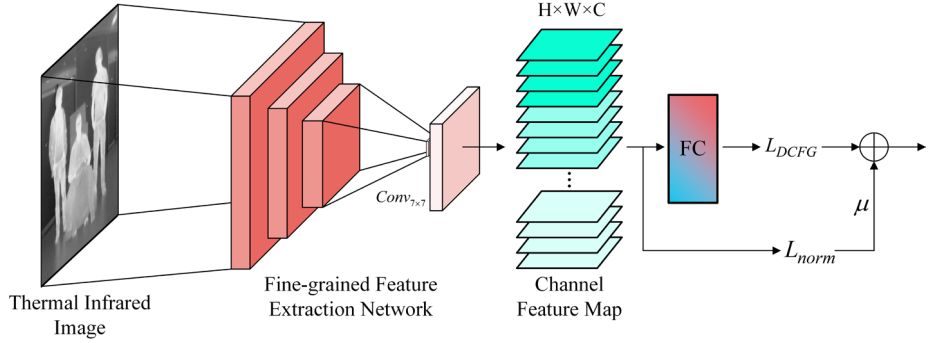

**Fig. 3.** Tracking loss function diagram.

**Diverse Cross-Channel Fine-Grained Loss Function.** To effectively extract diverse fine-grained TIR features, it is essential that different feature groups attend to distinct spatial regions of the target, avoiding redundancy in focusing on the same discriminative areas. In this work, we reduce the similarity between feature groups by introducing an inter-channel loss function, which maximizes feature diversity and encourages the tracker to learn a wide range of fine-grained representations. The overall structure of this loss function is illustrated in Fig. 4. The discriminative branch ensures that each feature channel contains meaningful discriminative information, while the diversity branch drives the network to focus on different target regions specific to the given category. The inter-channel loss function is designed to minimize the loss in the discriminative branch while maximizing the effectiveness of the diversity branch, ultimately improving the overall performance of the network.

The diverse cross-channel fine-grained loss $L_{DCFG}$ is defined as:

$$L_{DCFG} = \|coef(F_s) - I\|_2 \tag{13}$$

where $I$ is a diagonal matrix of size $(N, N)$, and $coef(\cdot)$ is a similarity function computed between elements of $f_s$, given by:

$$coef(F_s) = cov(f_{si}, f_{sj}) \times (\sigma f_{si} \sigma f_{sj})^{-1} \tag{14}$$

where $f_{si}$ and $f_{sj}$ represent different compressed features extracted from the cross-channel fine-grained feature blocks, with $i, j = 1,2,3,\ldots,N$.

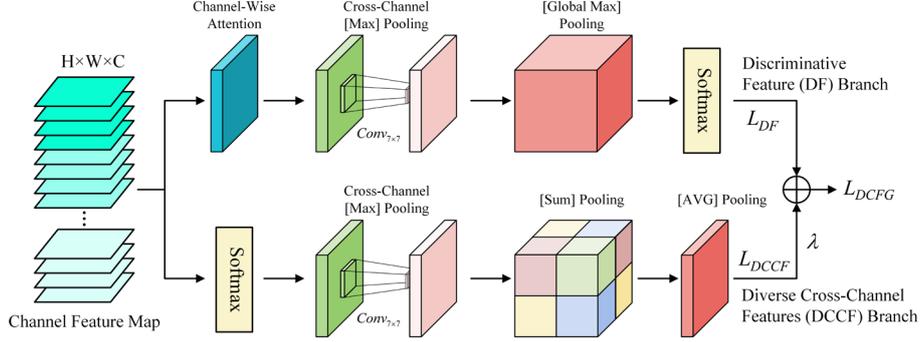

**Fig. 4.** Diverse cross-channel fine-grained loss function framework.

## 3      Experiment

To validate the effectiveness of the proposed DCFG, this section presents a comprehensive set of experimental results. Section 3.1 outlines the experimental setup. Section 3.2 reports ablation study results to analyze the contribution of each module. Section 3.3 evaluates the performance of DCFG on four TIR tracking benchmarks: LSOTB-TIR [12], PTB-TIR [13], VOT-TIR 2015 [14], and VOT-TIR 2017 [15]. Section 3.4 examines the robustness of the tracker under challenging conditions using the LSOTB-TIR and PTB-TIR datasets. Lastly, Section 3.5 provides visualizations of tracking outcomes to intuitively illustrate the effectiveness of the proposed tracker.

### 3.1    Implementation Details

**Experimental Environment.** The proposed DCFG tracker is implemented using Python 3.7 and PyTorch 1.8.1. The training process utilizes both visible-spectrum datasets: COCO and VOT2017, as well as the TIR dataset LSOTB-TIR. CUDA 11.8 and cuDNN 7.6 serve as the acceleration backends. All experiments are conducted on a workstation equipped with an Intel i7-13700KF CPU, 32 GB of RAM, and an NVIDIA GeForce RTX 3090 GPU with 24 GB of memory, running Ubuntu 20.04.

**Datasets and Evaluation Criteria.** LSOTB-TIR is currently the most comprehensive and diverse benchmark available for TIR target tracking. It comprises over 1,400 TIR video sequences spanning 47 object categories, with more than 600,000 frames and 730,000 manually annotated bounding boxes. The dataset includes four scene-level attributes and twelve challenge attributes, designed to support a thorough evaluation of TIR tracker performance. The PTB-TIR dataset contains 60 TIR pedestrian sequences and over 30,000 frames, with all annotations manually labeled to ensure evaluation accuracy. Following the protocols outlined in [12] and [13], Precision and Success are used as the primary evaluation metrics for the PTB-TIR benchmark. Additionally, Normalized Precision (NP) is adopted as a supplementary metric for LSOTB-TIR.

VOT-TIR2015, released by the Visual Object Tracking (VOT) committee, was the first benchmark tailored specifically for short-term TIR tracking. It includes 20 test sequences covering eight target categories, with an average sequence length of 563 frames. The VOT-TIR2017 dataset extends this with 25 test sequences, averaging 740 frames in length. In accordance with [14] and [15], we use Accuracy, Robustness, and Expected Average Overlap (EAO) as the evaluation metrics for both the VOT-TIR2015 and VOT-TIR2017 benchmarks.

### 3.2   Ablation Experiment and Analysis

Ablation experiments are performed on the PTB-TIR and LSOTB-TIR benchmarks to assess the individual contributions of the proposed components, namely the DCFG loss function, the DCCFG, and the Channel Balancing and Lightweight (CBL) strategy. The results of these experiments are summarized in Table 1, where SiamRPN++ is used as the baseline tracker and DCFG denotes the complete proposed tracking method.

**Table 1.** Ablation results on PTB-TIR and LSOTB-TIR benchmarks.

| Trackers | Component | | | PTB-TIR | | LSOTB-TIR | | |
|---|---|---|---|---|---|---|---|---|
| | DCCFG | CBL | DCFG Loss | Pre.↑ | Success↑ | Pre.↑ | Norm. Pre.↑ | Success↑ |
| SamRPN++(base) | | | | 0.712 | 0.582 | 0.740 | 0.692 | 0.554 |
| Siam_DCCFG | √ | | | 0.767 | 0.612 | 0.798 | 0.745 | 0.590 |
| Siam_CBL | | √ | | 0.777 | 0.629 | 0.812 | 0.750 | 0.599 |
| DCFG (ours) | √ | √ | √ | **0.841** | **0.639** | **0.849** | **0.766** | **0.616** |

Table 1 illustrates the individual contributions of each proposed component to the overall tracking performance. First, the Siam_DCCFG model surpasses the baseline SiamRPN++, yielding a 5.5% improvement in Precision and a 3.0% increase in Success rate on the PTB-TIR dataset. On the LSOTB-TIR benchmark, it achieves gains of 5.8% in precision, 5.3% in normalized precision, and 3.6% in Success rate. As Siam_DCCFG integrates only the DCCFG module into the baseline framework, these results underscore the strong impact of the proposed diverse cross-channel fine-grained feature learning network. Second, the Siam_CBL model enhances Precision by 6.5% and 7.2% on PTB-TIR and LSOTB-TIR, respectively, and increases Success rates by 4.7% and 4.5%. These findings validate the effectiveness of the channel balancing and lightweight design strategy. Finally, the complete DCFG tracker further outperforms Siam_CBL on both benchmarks, achieving an additional 6.4% and 3.7% gain in precision, along with 1.0% and 1.7% improvement in Success rate on PTB-TIR and LSOTB-TIR, respectively. These results further affirm the effectiveness of the proposed loss function in enhancing overall tracking performance.

To examine the influence of the number of feature groups (FGN) on the performance of the DCFG tracker, Table 2 reports the experimental results of the diverse cross-channel fine-grained feature learning network under various FGN settings. As the number of groups increases, both Precision and Success rate show consistent improvement. When FGN is set to 8, the DCFG tracker achieves performance gains of 6.7% and 7.6% in Precision and Success rate, respectively, on the LSOTB-TIR dataset, and 6.8% and 3.7% on the PTB-TIR dataset, compared to the baseline configuration.

However, further increasing the FGN to 16 leads to a noticeable decline in all evaluation metrics across both benchmarks. This performance drop is primarily due to the reduced representational capacity of each feature group, as the number of channels per group becomes insufficient. Specifically, with 256 total channels, dividing them into 16 groups results in only 16 channels per group. This configuration significantly increases both MAC and FLOPs, introducing computational overhead. As a result, when FGN exceeds a certain threshold, the overall tracking performance deteriorates [16]. Based on these observations, FGN = 8 is selected as the optimal configuration, striking a balance between fine-grained representation and computational efficiency.

**Table 2.** Performance comparison of different numbers of feature group N on LSOTB-TIR and PTB-TIR benchmark.

| Feature Group Number (FGN) | | $N=0$ | $N=1$ | $N=2$ | $N=4$ | $N=8$ | $N=16$ |
|---|---|---|---|---|---|---|---|
| LSOTB-TIR | Precision↑ | 0.782 | 0.792 | 0.806 | 0.845 | **0.849** | 0.830 |
| | Norm. Precision↑ | 0.733 | 0.746 | 0.750 | **0.767** | 0.766 | 0.757 |
| | Success↑ | 0.548 | 0.569 | 0.593 | 0.610 | **0.616** | 0.609 |
| PTB-TIR | Precision↑ | 0.765 | 0.777 | 0.799 | 0.830 | **0.841** | 0.824 |
| | Success↑ | 0.602 | 0.610 | 0.619 | 0.621 | **0.639** | 0.610 |

### 3.3   Performance Comparison with State-of-the-arts

The proposed DCFG tracker is evaluated against a wide range of state-of-the-art trackers on the LSOTB-TIR, PTB-TIR, VOT-TIR 2015, and VOT-TIR 2017 benchmarks. These trackers include correlation filter-based trackers, which are divided into traditional correlation filter approaches such as HCF [17], KCF [17], HDT [18], SRDCF [19], and STAPLE [20], as well as enhanced correlation filter approaches such as ECO [21], ECO-stir [22], ECO_LS [23], ECOHG_LS [23], ECO-deep [24], ECO-MM [25], DeepSTRCF, MCFTS [26], MCCT [27], and UDCT [28]. In addition, a number of Siamese network-based trackers are considered. These include original Siamese frameworks such as CFNet [29], SiamFC [30], DaSiamRPN [31], SiamRPN [31], SiamRPN++ [32], SiamMask [33], and SiamSAV [34], as well as more advanced Siamese models with enhanced functionalities, including Ocean [35], TADT [36], MMNet [37], HSSNet [38], and MLSSNet [39]. Deep learning-based trackers are also included in the evaluation, such as AMFT [40], MDNet [41], VITAL [42], CREST [43], ATOM [44], GFSNet [45], and DiMP [46]. Furthermore, the evaluation includes Transformer-based trackers such as TransT [47], CSWinTT [48], DFG [49], MixFormer [50], and CorrFormer [51]. The experimental results are summarized in Fig. 5, Fig. 6, and Table 3.

**Results on PTB-TIR.** As illustrated in Fig. 5, our DCFG achieves the highest performance in both Precision and Success rate, reaching 0.841 and 0.639, respectively. Compared with the baseline SiamRPN++, the DCFG shows improvements of 9.0% in Precision and 10.3% in Success rate. Relative to correlation filter-based trackers such as ECO-stir and MCFTS, it achieves gains of 1.2% and 13.2% in Success rate, respectively. Moreover, the DCFG outperforms Transformer-based trackers such as DFG and TransT in Precision by 1.6% and 6.1%, respectively. These results demonstrate that leveraging cross-channel fine-grained features enables the DCFG tracker to generate more accurate and robust representations of TIR targets, leading to better performance in both Precision and Success rate compared to existing TIR tracking approaches.

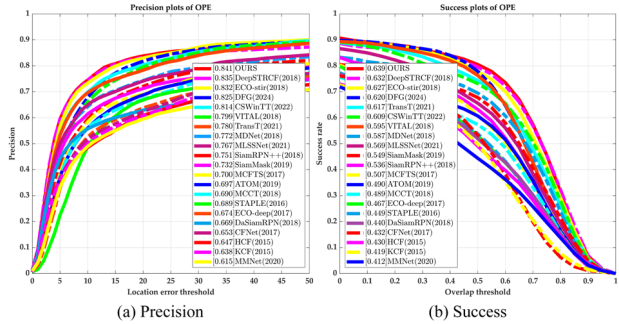

**Fig. 5.** Results on PTB-TIR benchmark.

**Results on LSOTB-TIR.** As illustrated in Fig. 6, our DCFG achieves the highest Precision score of 0.849, outperforming both AMFT and DFG. Compared to TransT, DCFG demonstrates improvements of 2.1% in Precision and 1.2% in Success rate. Additionally, it attains the highest Success rate of 0.616. When evaluated against Siamese network-based trackers such as SiamRPN++ and MLSSNet, DCFG achieves Success rate gains of 6.2% and 8.1%, respectively. These results confirm that DCFG outperforms competing approaches on the LSOTB-TIR benchmark, which can be attributed to the effectiveness of its diverse cross-channel fine-grained feature learning network.

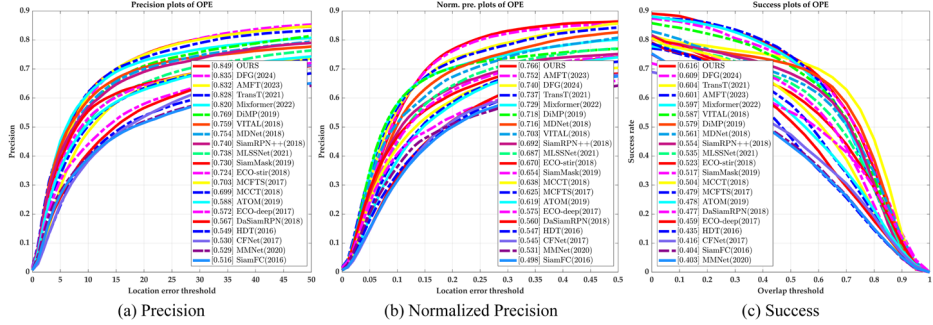

(a) Precision    (b) Normalized Precision    (c) Success

**Fig. 6.** Results on LSOTB-TIR benchmark.

**Results on VOT-TIR2015 and VOT-TIR2017.** As shown in Table 3, the proposed DCFG tracker achieves the highest accuracy scores on both the VOT-TIR2015 and VOT-TIR2017 benchmarks, reaching 0.81 and 0.78, respectively, and outperforming trackers such as GFSNet, Ocean, and ECO-MM. In addition, DCFG attains the best EAO score of 0.356 on the VOT-TIR2017 benchmark, and an EAO of 0.384 on VOT-TIR2015, ranking second only to the Siamese network-based tracker UDCT. Compared with the baseline SiamRPN++, the DCFG improves accuracy by 7.0% and 9.0% on VOT-TIR2015 and VOT-TIR2017, respectively, while achieving EAO gains of 7.1% and 6.0%. Lastly, the DCFG tracker achieves a robustness score of 1.45, which is lower than those of UDCT, VITAL, and ECO-MM. These results demonstrate that the proposed diverse cross-channel fine-grained feature learning network substantially improves the tracker's ability to distinguish targets from distractors.

**Table 3.** Results on VOT-TIR 2015 and VOT-TIR 2017 benchmarks.

| Methods | Trackers | Year | VOT-TIR 2015 | | | VOT-TIR 2017 | | |
|---|---|---|---|---|---|---|---|---|
| | | | EAO↑ | Acc↑ | Rob↓ | EAO↑ | Acc↑ | Rob↓ |
| Correlation filter | SRDCF | 2015 | 0.225 | 0.62 | 3.06 | 0.197 | 0.59 | 3.84 |
| | HDT | 2016 | 0.188 | 0.53 | 5.22 | 0.196 | 0.51 | 4.93 |
| | ECO-deep | 2017 | 0.286 | 0.64 | 2.36 | 0.267 | 0.61 | 2.73 |
| | MCFTS | 2017 | 0.218 | 0.59 | 4.12 | 0.193 | 0.55 | 4.72 |
| | MCCT | 2018 | 0.250 | 0.67 | 3.34 | 0.270 | 0.53 | 1.76 |
| | ECO-MM | 2022 | 0.303 | 0.72 | 2.44 | 0.291 | 0.65 | 2.31 |
| | ECO_LS | 2023 | 0.319 | 0.64 | 0.82 | 0.302 | 0.55 | 0.93 |
| | ECOHG_LS | 2023 | 0.270 | 0.60 | 0.92 | 0.251 | 0.49 | 1.26 |
| Transformer | TransT | 2021 | 0.287 | 0.77 | 2.75 | 0.290 | 0.71 | 0.69 |
| | DFG | 2024 | 0.329 | 0.78 | 2.41 | 0.304 | 0.74 | 2.63 |
| | CorrFormer | 2023 | 0.269 | 0.71 | 0.56 | 0.262 | 0.66 | 1.23 |
| Deep learning | CREST | 2017 | 0.258 | 0.62 | 3.11 | 0.252 | 0.59 | 3.26 |
| | DeepSTRCF | 2018 | 0.257 | 0.63 | 2.93 | 0.262 | 0.62 | 3.32 |
| | VITAL | 2018 | 0.289 | 0.63 | 2.18 | 0.272 | 0.64 | 2.68 |
| | DiMP | 2019 | 0.330 | 0.69 | 2.23 | 0.328 | 0.66 | 2.38 |
| | ATOM | 2019 | 0.331 | 0.65 | 2.24 | 0.290 | 0.61 | 2.43 |
| | Ocean | 2020 | 0.339 | 0.70 | 2.43 | 0.320 | 0.68 | 2.83 |
| | SiamFC | 2016 | 0.219 | 0.60 | 4.10 | 0.188 | 0.50 | **0.59** |
| | CFNet | 2017 | 0.282 | 0.55 | 2.82 | 0.254 | 0.52 | 3.45 |
| | DaSiamRPN | 2018 | 0.311 | 0.67 | 2.33 | 0.258 | 0.62 | 2.90 |
| | SiamRPN | 2018 | 0.267 | 0.63 | 2.53 | 0.242 | 0.60 | 3.19 |
| | TADT | 2019 | 0.234 | 0.61 | 3.33 | 0.262 | 0.60 | 3.18 |
| | HSSNet | 2019 | 0.311 | 0.67 | 2.53 | 0.262 | 0.58 | 3.33 |
| Siamese network | SiamRPN++ | 2019 | 0.313 | 0.74 | 2.25 | 0.296 | 0.69 | 2.63 |
| | MMNet | 2020 | 0.340 | 0.61 | 2.09 | 0.320 | 0.58 | 2.90 |
| | MLSSNet | 2020 | 0.329 | 0.57 | 2.42 | 0.286 | 0.56 | 3.11 |
| | SiamSAV | 2021 | 0.344 | 0.71 | 2.75 | 0.336 | 0.67 | 2.92 |
| | UDCT | 2022 | **0.420** | 0.67 | 0.88 | 0.342 | 0.66 | 0.81 |
| | GFSNet | 2022 | 0.370 | 0.67 | **0.21** | 0.354 | 0.62 | 0.64 |
| | DCFG (ours) | - | 0.384 | **0.81** | 1.45 | **0.356** | **0.78** | 1.87 |

### 3.4   Tracking Challenges: Results and Analysis

Experiments were conducted on the twelve predefined evaluation attributes within the PTB-TIR and LSOTB-TIR benchmarks, and performance comparisons were made against several trackers on these datasets. These evaluation attributes encompass a broad range of challenging tracking conditions, including Deformation (DF), Occlusion (OCC), Scale Variation (SV), Background Clutter (BC), Low Resolution (LR), Fast Motion (FM), Motion Blur (MB), Out of View (OV), Intensity Variation (IV), Thermal Crossover (TC), Aspect Ratio Change (ARC), and Distractors (DI). The corresponding results are reported in Table 4 and Table 5.

**Tracking Challenge Results on the PTB-TIR Benchmark.** As shown in Table 4, the proposed DCFG tracker achieves the best overall performance across all 12 evaluation attributes on the PTB-TIR benchmark, with a Precision of 82.2% and a Success rate of 61.5%. Furthermore, DCFG delivers better performance on 6 of the 12 attributes, outperforming strong baselines such as TransT and MDNet. Although DeepSTRCF achieves the top result under the OCC attribute, its average Precision remains 3.2% lower than that of the proposed tracker. Notably, DCFG records the highest Precision and Success rate under the DI attribute, reaching 81.8% and 42.5%, respectively. This strong performance is attributed to the effectiveness of the proposed diverse cross-channel fine-grained feature learning network in extracting highly discriminative target features, which enables the tracker to better distinguish the target from surrounding interference. Overall, the experimental results demonstrate that DCFG exhibits significant advantages in addressing tracking challenges involving TIR pedestrian targets—particularly under difficult conditions such as DI, DF, OCC, LR, MB, and Intensity IV.

Table 4. The results of tracking challenges on the PTB-TIR benchmark.

| Attributes | Tracker (P/%↑ S/%↑) | | | | | | |
|---|---|---|---|---|---|---|---|
| | Deep-STRCF | MDNet | VITAL | MDNet | TransT | ECO-stir | DCFG |
| DF  | 80.8 / 43.5 | 70.8 / 41.8 | 76.3 / 51.7 | 78.5 / 51.5 | 82.3 / 32.5 | 84.7 / 40.4 | **87.2 / 54.9** |
| OCC | **83.2 / 68.6** | 64.5 / 53.1 | 69.2 / 58.0 | 70.4 / 44.5 | 70.3 / 51.0 | 72.9 / 48.9 | 77.2 / 65.6 |
| SV  | 86.7 / 61.0 | 79.1 / 57.1 | 87.1 / **71.5** | 60.2 / 39.5 | 79.6 / 59.7 | **94.5** / 67.3 | 94.4 / 71.0 |
| BC  | 75.1 / **65.2** | 71.8 / 50.3 | 74.5 / 58.5 | **86.3** / 50.2 | 75.3 / 55.3 | 74.9 / 55.3 | 67.5 / 52.5 |
| LR  | 78.7 / 57.0 | 89.7 / 55.8 | 85.2 / 62.5 | 80.6 / 44.4 | 91.4 / 56.1 | 76.9 / 54.3 | **96.6 / 64.7** |
| FM  | 81.9 / 58.8 | 82.7 / 53.1 | 76.9 / 69.0 | 68.1 / 58.0 | 70.9 / 58.9 | 81.5 / 71.5 | **82.8 / 70.0** |
| MB  | 70.2 / 55.0 | 75.1 / 52.4 | **89.2 / 69.0** | 76.4 / 40.4 | 75.8 / 52.6 | 84.9 / 56.3 | 81.3 / 66.0 |
| OV  | 79.5 / **71.9** | **86.3** / 55.5 | 68.7 / 64.0 | 71.5 / 40.0 | 77.1 / 50.5 | 78.9 / 60.8 | 85.8 / 63.0 |
| IV  | 87.3 / 72.0 | 82.5 / 56.5 | 86.9 / 70.7 | 81.1 / 63.5 | 77.6 / 66.5 | 87.2 / 66.6 | **90.3 / 74.6** |
| TC  | **77.8** / 41.0 | 65.6 / **54.1** | 73.6 / 47.0 | 70.7 / 34.0 | 72.8 / 38.2 | 55.1 / 49.6 | 63.4 / 47.6 |
| ARC | 75.3 / 57.5 | 68.3 / 59.0 | 68.2 / 59.0 | 80.9 / 56.5 | 72.6 / 37.0 | 77.4 / 63.2 | **78.9 / 65.5** |
| DI  | 71.5 / 36.5 | 79.7 / 30.5 | 75.3 / 38.7 | 72.5 / 29.8 | 79.9 / 32.9 | 76.8 / 30.6 | **81.8 / 42.5** |
| ALL | 79.0 / 57.3 | 76.3 / 51.6 | 77.6 / 60.0 | 74.8 / 46.0 | 77.1 / 49.3 | 78.8 / 55.4 | **82.2 / 61.5** |

**Tracking Challenge Results on the LSOTB-TIR Benchmark.** As shown in Table 5, the DCFG achieves the highest scores across all evaluation metrics on the LSOTB-TIR benchmark, with a Precision of 79.1%, a normalized Precision of 73.4%, and a Success rate of 58.8%. Moreover, DCFG significantly outperforms other trackers on 7 out of the 12 evaluation attributes. Notably, compared with SiamRPN++, DCFG achieves improvements of 9.3% in precision, 16.0% in normalized precision, and 13.1% in Success rate. Although ATOM records the highest Precision under the BC attribute, DCFG still surpasses it in overall performance, with gains of 2.6% in precision, 9.3% in normalized precision, and 9.0% in Success rate. Additionally, the DCFG tracker continues to demonstrate strong performance under the DI attribute, achieving a Precision of 79.8%, a normalized Precision of 73.4%, and a Success rate of 58.8%. These results collectively highlight the robustness and effectiveness of the DCFG tracker in handling thermal infrared target tracking under complex and challenging scenarios.

Table 5. Attributes analysis on LSOTB-TIR benchmark.

| Attributes | Tracker (P/%↑ NP/%↑ S/%↑) | | | | | |
|---|---|---|---|---|---|---|
| | MDNet | SiamRPN++ | ATOM | SiamMask | ECO-stir | DCFG |
| DF  | 76.2/64.5/49.4 | 74.8/60.1/37.6 | 77.9/57.0/47.2 | 73.9/66.2/53.3 | 82.9/68.4/56.8 | **83.5/72.6/60.6** |
| OCC | 71.1/62.7/44.9 | 72.7/61.4/46.3 | 70.3/55.9/49.0 | **75.3/74.9**/52.2 | 74.3/66.3/53.2 | 73.6/69.4/**53.9** |
| SV  | **94.1/72.3**/33.3 | 64.3/58.6/44.9 | 78.6/66.7/52.5 | 84.2/77.2/65.1 | 88.0/76.5/**66.7** | 93.1/65.2/52.2 |
| BC  | 73.9/63.3/62.3 | 70.8/58.7/43.9 | **78.6/75.2/57.2** | 73.5/67.5/49.1 | 75.8/65.1/57.0 | 69.5/67.8/54.8 |
| LR  | 77.5/75.6/70.3 | 81.1/63.1/**52.3** | 93.7/67.9/49.5 | 82.6/65.3/45.3 | 75.5/67.4/46.5 | **89.8/79.8/66.2** |
| FM  | 73.8/71.9/64.5 | 72.2/54.6/53.8 | 72.6/65.4/47.9 | 74.6/65.5/58.7 | 84.6/75.3/64.1 | **85.9/79.5/65.1** |
| MB  | 76.7/69.3/60.3 | 76.8/59.9/47.8 | 78.4/56.3/50.3 | 75.7/62.5/57.1 | 83.1/74.3/61.5 | **85.6/75.6/63.9** |
| OV  | **76.8/72.3**/57.8 | 63.8/60.6/45.7 | 69.1/65.5/48.3 | 73.6/66.1/58.7 | 75.4/68.6/**59.6** | 71.2/69.8/58.5 |
| IV | 79.0/76.9/75.4 | 68.4/64.1/61.7 | 83.6/71.4/53.4 | 84.5/73.9/72.7 | 87.7/82.7/69.8 | **89.5/87.9/71.3** |
| TC | **73.1**/**70.3**/41.5 | 58.0/47.6/37.3 | 70.8/54.6/48.7 | 58.6/49.0/40.2 | 55.3/52.0/**42.9** | 64.7/66.8/42.5 |
| ARC | 72.4/62.1/61.5 | 72.9/49.1/32.1 | 70.1/60.4/46.7 | 70.2/48.4/56.4 | 78.6/64.5/54.5 | **80.4/73.5/58.9** |
| DI | 74.7/64.9/59.6 | 61.6/51.4/44.6 | 73.8/72.9/46.6 | 71.6/64.9/48.9 | 73.6/65.6/53.7 | **79.8/77.6/57.2** |
| ALL | 76.6/68.8/49.2 | 69.8/57.4/45.7 | 76.5/64.1/49.8 | 74.9/65.1/54.8 | 77.9/68.9/57.2 | **79.1/73.4/58.8** |

### 3.5 Visualized Comparison Results

Fig 7 illustrates the visual tracking performance of the proposed DCFG tracker in comparison with AMFT and DeepSTRCF across five sequences from the PTB-TIR dataset, where the ground truth represents the annotated bounding box of the target. As shown in Fig 7(a), 7(b), and 7(d), both AMFT and DeepSTRCF fail to maintain accurate tracking when distractor objects appear in the scene. In contrast, the DCFG tracker effectively distinguishes the target from distractors, maintaining accurate localization throughout the sequences. In frame 350 of Fig 7(c), the target becomes occluded, causing AMFT and DeepSTRCF to lose track. However, leveraging its diverse cross-channel fine-grained feature representation, the DCFG tracker exhibits strong robustness and continues to track the occluded target accurately.

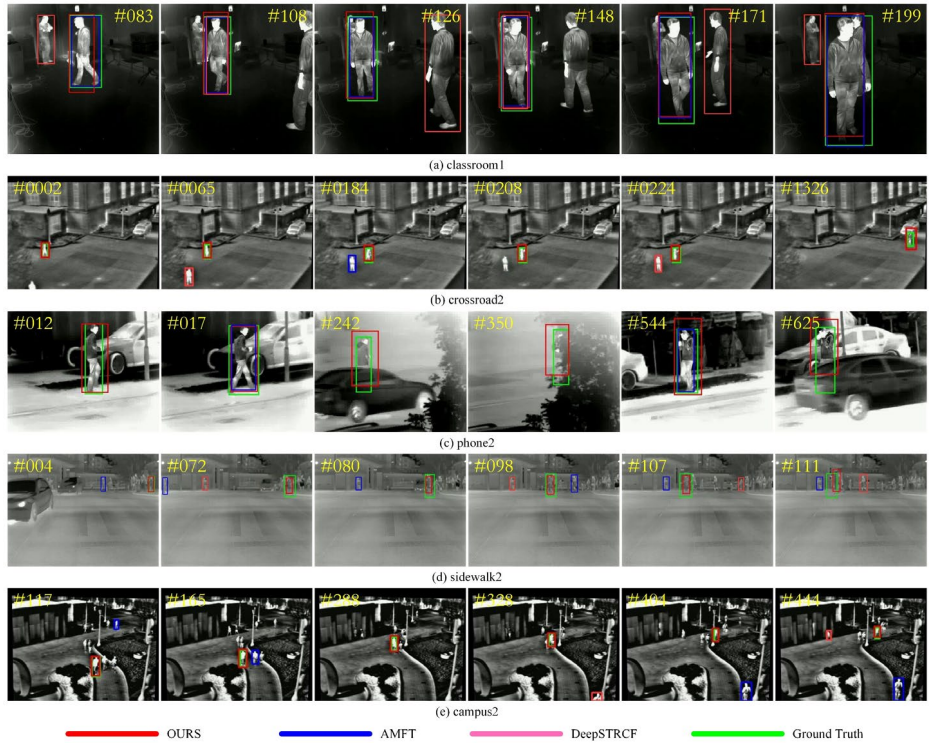

**Fig. 7.** Visualization results of the qualitative comparison experiments conducted on six challenging sequences in the PTB-TIR and LSOTB-TIR benchmarks.

Furthermore, Fig 7(e) demonstrates the reliable performance of the DCFG tracker under challenging conditions such as background blur, low resolution, and thermal crossover. In these scenarios, both comparison trackers either lose the target or produce inaccurate tracking results. These visual comparisons confirm the superior tracking capability of the proposed DCFG tracker over existing TIR trackers.

## 4    Conclusions

This paper proposes a TIR target tracking algorithm that leverages cross-channel fine-grained feature learning and progressive fusion. The proposed tracker comprises a diverse cross-channel fine-grained feature learning network and a specially designed loss function. The network suppresses salient target features through the use of masks and suppression coefficients, thereby facilitating the extraction of more informative fine-grained representations. Furthermore, the integration of channel rearrangement and channel equalization strategies optimizes information flow, reduces parameter overhead, and enhances the efficiency and compactness of the model. Feature redirection and channel shuffle mechanisms are also employed to enable effective fusion of fine-grained features. The loss function incorporates an inter-channel loss term that enforces near-orthogonality among TIR features across channels. This promotes feature diversity and strengthens the ability of network to capture fine-grained representations. Experimental results show that the proposed tracker achieves state-of-the-art performance, with a Precision of 0.849 and a Success rate of 0.616 on the LSOTB-TIR and PTB-TIR benchmarks, respectively. It also achieves top accuracy of 0.81 and an EAO of 0.356 on the VOT-TIR 2015 and VOT-TIR 2017 benchmarks, along with an accuracy of 0.78 on VOT-TIR 2017. In future work, we plan to explore the integration of interleaved group convolution and the design of lightweight, attention-based feature extraction modules. These enhancements aim to adaptively weight fine-grained features across channels, prioritize those that contribute more discriminative information, improve the quality of feature representations, and suppress background interference.